# BROOKINGS

*Center on Regulation and Markets Working Paper #9*

September 2023

# Market concentration implications of foundation models:
## *THE INVISIBLE HAND OF CHATGPT*

Jai Vipra and Anton Korinek

## Disclosure

Microsoft, Google, Meta, and Amazon provide support to The Brookings Institution. The findings, interpretations, and conclusions in this report are not influenced by any donation. Brookings recognizes that the value it provides is in its absolute commitment to quality, independence, and impact. Activities supported by its donors reflect this commitment.

Jai Vipra was supported by a fellowship with the Centre for the Governance of AI. Other than the aforementioned, the authors did not receive financial support from any firm or person for this article or from any firm or person with a financial or political interest in this article. The authors are not currently an officer, director, or board member of any organization with a financial or political interest in this article.

The Brookings Institution is financed through the support of a diverse array of foundations, corporations, governments, individuals, as well as an endowment. A list of donors can be found in our annual reports published online [here](). The findings, interpretations, and conclusions in this report are solely those of its author(s) and are not influenced by any donation.

# Market Concentration Implications of Foundation Models: The Invisible Hand of ChatGPT[1]

Jai Vipra (GovAI) and Anton Korinek (Brookings, UVA and GovAI)

We analyze the structure of the market for foundation models, i.e., large AI models such as those that power ChatGPT and that are adaptable to downstream uses, and we examine the implications for competition policy and regulation. We observe that the most capable models will have a tendency towards natural monopoly and may have potentially vast markets. This calls for a two-pronged regulatory response: (i) Antitrust authorities need to ensure the contestability of the market by tackling strategic behavior, in particular by ensuring that monopolies do not propagate vertically to downstream uses, and (ii) given the diminished potential for market discipline, there is a role for regulators to ensure that the most capable models meet sufficient quality standards (including safety, privacy, non-discrimination, reliability and interoperability standards) to maximally contribute to social welfare. Regulators should also ensure a level regulatory playing field between AI and non-AI applications in all sectors of the economy. For models that are behind the frontier, we expect competition to be quite intense, implying a more limited role for competition policy, although a role for regulation remains.

[1] We thank Emma Bluemke, Aidan Kane, Sarah Myers West, Sanjay Patnaik, Nicholas Ritter, Max Schnidman, Eli Schrag and Rob Seamans for their thoughtful comments. Any remaining errors are our own.

## Executive Summary

Foundation models are large artificial intelligence (AI) models that can be adapted for use in a wide range of downstream applications. As foundation models grow increasingly capable, they become useful for applications across a wide range of economic functions and industries. Ultimately, the potential market for foundation models may encompass the entire economy. This implies that the stakes for competition policy are tremendous.

We find that the market for cutting-edge foundation models exhibits a strong tendency towards market concentration: The fixed costs of training a foundation model are high, and the marginal cost of deploying them are very low. This means that there are large economies of scale in operating foundation models—the average cost of producing one unit of output declines the greater the scale of deployment. There are also some economies of scope—it is cheaper for one AI company to produce multiple foundation models for different uses than for multiple AI companies to cater to these uses separately. First-mover advantages in the market for foundation models are high, although they require large ongoing investments in product deployment, marketing, and distribution. Other barriers to entry, such as limited resources like talent, data, computational power, and intellectual property protections also create forces that point towards natural monopoly.

One particular concern for competition policy is that the producers of foundation models could expand their market power vertically to downstream uses, in which competition would otherwise ensure lower prices and better services for users. Producers may also erect barriers to entry or engage in predatory pricing, which may make the market for foundation models less contestable. The negative implications of excessive concentration and lack of contestability in the market for foundation models include the standard monopoly distortions, ranging from restricted supply and higher prices to the resulting implications for the concentration of economic power and inequality. Moreover, they may include the systemic risks and vulnerabilities that arise if a single model or small set of models are deployed extensively throughout the economy, and they may give rise to growing regulatory capture. On the other hand, concentration in the market for foundation models may allow producers to better internalize any potential safety risks of such systems, including the risks of accidents and malicious use since competitive pressures might induce producers to deploy AI products more quickly and invest less in safety research. The rise of open-source models mitigates concerns about market concentration but comes with its own set of downsides, including growing safety risks and the potential for abuse by malicious actors.

We conclude that regulators are well-advised to adopt a two-pronged strategy in response to these economic and safety factors:

First, as the market for the most capable foundation models has characteristics that point towards market concentration, it is important to ensure that it remains contestable and that incumbents do not engage in strategic behavior to deter innovation and the entry of new firms, e.g., via predatory pricing or strategic lobbying. Regulators must pay particular attention to risks arising from vertical integration with both upstream and downstream producers. The equilibrium market structure may be a single, or a small number of, producers of leading foundation models, which would enable many actors across different sectors of the economy to fine-tune or deploy their models in downstream applications. Attention also needs to be paid to the market for downstream applications to remain competitive.

Second, as natural monopolies or oligopolies, producers of the most advanced foundation models may need to be regulated akin to public utilities. Since market forces are blunted in the presence of market concentration, regulators need to ensure that users experience reasonable pricing, high quality standards (including safety, privacy, non-discrimination, reliability, and interoperability standards), as well as disclosure and equal access rights. Moreover, the regulators of foundation models need to recognize the growing systemic importance of these systems as they are deployed in increasingly important roles throughout our economy.

Regulators should also ensure that AI products and services compete on a level playing field with non-AI products and services, including human-provided services. Sectoral regulations on liability, professional licensing, and professional ethics should apply equally as is appropriate to both AI and non-AI solutions. For instance, hiring decisions and credit decisions must be subject to the same rules against discrimination and bias, no matter whether they are made by humans or AI. Likewise, financial advice should be subject to similar kinds of regulation regardless of whether it is provided by humans or AI. This may reduce AI use in some sectors, and simultaneously avoid the degradation of service standards through the use of AI.

## Table of Contents

# Introduction

In a post titled "Moore's Law for Everything", OpenAI CEO Sam Altman predicted that within the next few decades, AI technology would "do almost everything, including making new scientific discoveries that will expand our concept of 'everything'" (Altman 2021). Another recent paper co-authored by researchers at OpenAI estimated that most occupations are exposed to the deployment of large language models (LLMs) like ChatGPT, and that once complementary investments are made, up to 49 percent of workers could have half or more of their tasks exposed to LLMs (Eloundou et al. 2023). OpenAI itself has called for greater regulation of AI (Kang 2023). The fact that a private entity predicts that its commercial offering may cause widespread economic disruption – and consequently accumulation of economic power – underlines the need for policymakers, including antitrust authorities, to pay close attention.

LLMs belong to a class of models called foundation models – large artificial intelligence models that use machine learning methods, are trained on vast amounts of data, and can be adapted to various tasks and applications (Bommasani et al. 2021). The category of foundation models also includes other generative AI models such as vision models like the Vision Transformer (Dosovitskiy et al. 2021), audio models like Whisper (Radford et al. 2022), and multimodal models like CLIP (Radford et al. 2021) and Gato (Reed et al. 2022).

Foundation models are trained at a great cost, consisting primarily of the cost of computational power, data, and talent. The first step in the training of such models is called "pretraining" and delivers the foundation model itself. The process of adapting a foundation model to downstream tasks is called "fine-tuning" (Bommasani et al. 2021). This involves training the model on (generally) labeled and task-specific datasets.

Foundation models have shown great promise in performing economically useful tasks, sometimes much faster and at lower cost than human workers. They have been able to generate images according to broad instructions, write code, find errors in code, write essays, summarize documents, generate ideas, recognize speech, and so on.

This article analyzes the market concentration implications of foundation models. We believe that this is important for several reasons. The most advanced models currently are being developed in a handful of private entities, all of which are linked to existing Big Tech companies. The broad scope of use of these models implies wide-ranging impacts on the economy and society. Because foundation models can affect so many economic tasks, any tendency towards monopolization may mean that incumbents do not compete just *in a single market* or even *for a single market*: in the limit, they may compete for the entire economy. In other words, since their capabilities are so broad and improving so quickly, it is possible that foundation models could play a significant role in all production and exchange relationships in the economy. While foundation models are yet to be embedded in large-scale economic functions, competition regulators around the world have acknowledged the importance of carefully monitoring the sector (Khan 2023).

We analyze the structure of the market for foundation models as commercial products, and the potential barriers to entry to the market of foundation models. We describe important congruences and departures from the market concentration created by platforms and recommend policy measures that can be taken to create the conditions for a competitive market. We also discuss that the natural monopoly dynamics of foundation models create a role for utility-style regulation. Moreover, we observe the risk that competition for the market may push the producers of foundation models towards dangerously competitive races where unsafe models are rapidly released, and we analyze appropriate guardrails.

A fulcrum of the public discussion on the contestability of the market for foundation models was a leaked memo from a senior engineer at Google in May 2023.[2] This memo claimed that Google and OpenAI operated in a market without any significant barriers to entry, and that "owning the open-source ecosystem"[3] was key to success in this market. By contrast, we show in this article that there are indeed significant barriers to entry to the market for the most cutting-edge foundation models. Moreover, we also recognize that incumbents are currently incentivized to vertically integrate – including by dominating the open-source ecosystem at various points of the production chain – to cement their market position.

The central assumption of this article is that cutting-edge foundation models have reached the point where the output that they produce is economically useful and that their usefulness is going to increase in the coming years.[4]

There is some early evidence for the economic usefulness of foundation models (Baily, Brynjolfsson, and Korinek 2023). Every day, use cases emerge and transform. Software engineers, writers, and even economists can find their productivity improved significantly using foundation models. Google and Microsoft have integrated foundation models into their own products.

Additionally, a Goldman Sachs report finds that generative AI could increase annual global GDP by 7 percent (Hatzius et al. 2023). There are also reports that foundation models are already displacing video game illustrator jobs in China (Zhou 2023). A survey of 1,000 US business leaders found that one quarter had replaced some workers with ChatGPT, and one quarter had saved USD 50,000 or more by using ChatGPT (ResumeBuilder.com 2023). When Chegg, an education services company, reported that ChatGPT was impacting its business, it subsequently experienced a 48 percent decline in its share price (Thomson 2023).

---

[2] See https://www.semianalysis.com/p/google-we-have-no-moat-and-neither for the text of the memo.

[3] We use "open-source ecosystem" to refer to elements related to the development, distribution, and use of open-source software in a given sector. These elements include projects, collaborative tools, community forums, version control and hosting platforms, license systems and governance. A firm with a dominant position in this may be able to derive a strategic advantage over its competitors by attracting more talent, influencing the direction of the sector, creating a loyal user base, etc.

[4] Although the tendency of language models to "hallucinate," i.e., to produce factually incorrect results is concerning, cutting-edge models have shown significant improvements in this regard.

**Principal players in the market for foundation models**

The current market for foundation models is characterized by significant concentration. There are a few notable producers of frontier foundation models who each have at least some degree of profit motives and several open-source challengers. Figure 1 illustrates the market share of leading commercial LLM chatbots.

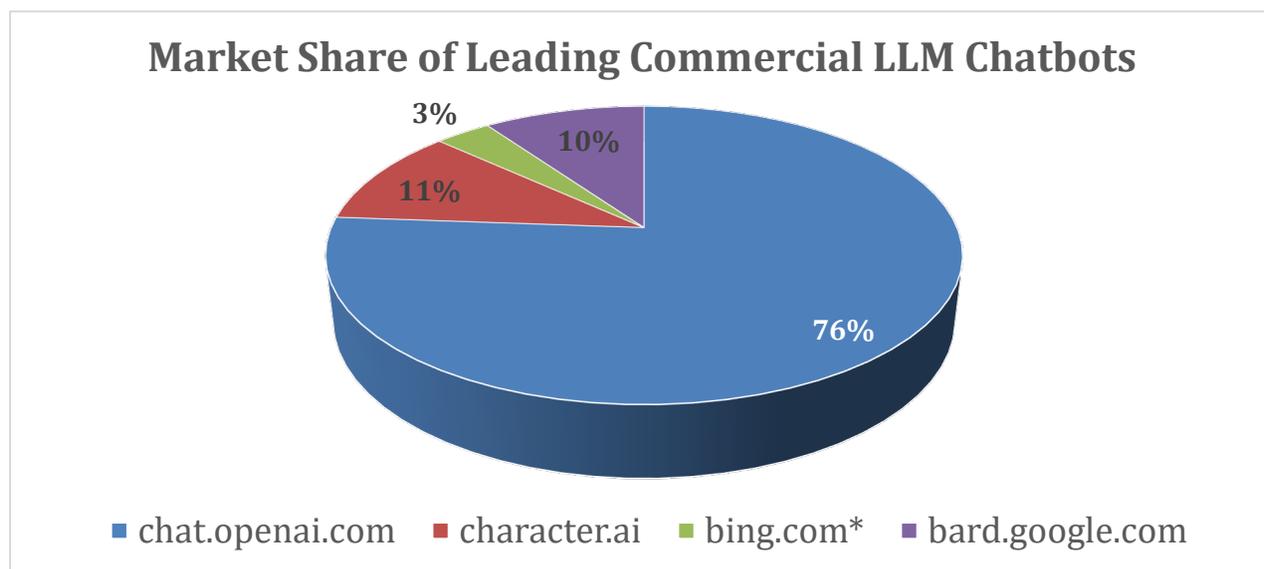

Figure 1: Fraction of estimated total visits in July 2023. Source: Authors' calculations based on similarweb.com. *Data for bing.com is estimated based on traffic increase to bing.com since January 2023, the month before the launch of the New Bing chatbot.

OpenAI, the producer of ChatGPT, is the clear market leader as of mid-2023, given the cutting-edge technology underlying its GPT-4 model and the commercial success of its ChatGPT interface. OpenAI is structured as a for-profit entity that is majority-owned by the OpenAI Nonprofit, to which all profits go once outside investors (such as Microsoft) have received profits up to an agreed multiple of their initial investment.[5] This profit cap, while unusual, is still far from being reached, implying that investors still have a substantial upside in the medium term, giving OpenAI a profit motive.[6]

---

[5] The board of the non-profit entity is allowed to have only a minority of members who own financial stakes in the for-profit entity, and important decision-making powers are limited to members who do not own such stakes. For more, see "Our Structure" on the OpenAI website https://openai.com/our-structure.

[6] The profit cap is negotiated separately for each investor. For Microsoft's USD 10 billion investment, this profit cap is reportedly equal to 100 times the initial investment (Coldewey 2019). For comparison, the top one percentile of US venture capital funds generate a total value to paid-in capital ratio of around 29, and a distribution to paid-in capital ratio of around 22. For more, see https://www.british-business-bank.co.uk/uk-venture-capital-financial-returns-2022/.

The second player in the market with comparable offerings is Alphabet's Google DeepMind unit, which is a result of a 2023 merger between Google Brain, Google's internal frontier AI department, and DeepMind, which Google acquired in 2014 (Shu 2014). DeepMind has developed seminal AI models such as AlphaGo and AlphaFold, while Google Brain researchers developed the Transformer architecture upon which today's LLMs are based. Google DeepMind's latest foundation model is PaLM 2, which has capabilities that come close to GPT-4 (Elias 2023). Another player is Anthropic, known for its LLM-based chatbot Claude, and more recently, Claude 2. Anthropic has significant external funding from Alphabet (Hu and Shekhawat 2023). Figure 2 shows that there is also significant concentration when leading LLM labs are measured by their valuation.

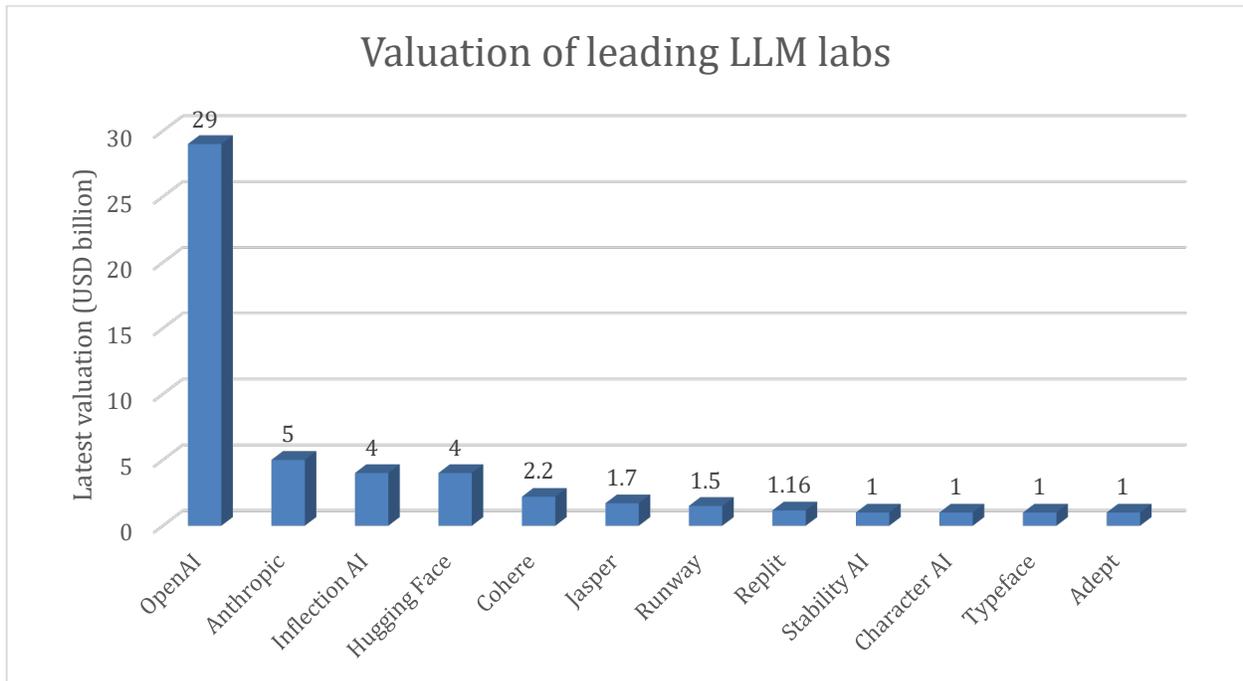

Figure 2: Most recent valuation as of July 2023. Source: Data collected by authors through news reports.

The market for models behind the frontier is dynamic and competitive and includes many open-source models. Meta released LlaMA (Touvron et al. 2023), an LLM that was made available to outside researchers who have built a large number of fine-tuned versions of it. Its successor, LlaMA 2, is a suite of models ranging from 7 billion to 70 billion parameters that Meta has made available for free for both research and commercial uses in July 2023 (Meta 2023). Moreover, Meta has released a code generation model (Code LlaMa) and announced a model that can handle multiple types of data inputs (text, audio, video, images, movements, etc.) and that is also planned to be open-sourced for both research and commercial uses (O'Regan, Victor, and Efrati 2023). BLOOM is a notable early example of a Transformer-based open-source foundation model developed collaboratively by a large number of researchers (Le Scao et al. 2023). Some models like Alpaca are open-source but trained on the outputs of closed models. Other examples of leading open-source models are Falcon (Penedo et al. 2023),which outperforms Meta's LlaMA in some benchmarks, and Guanaco (Dettmers et al. 2023), which improves upon LlaMA in some ways and is open for research use. However, the capabilities of open-source models today tend to lag behind the most powerful commercial models described above.

Additionally, it is important to closely examine the deep involvement of large technology companies in the market for foundation models.. For example, Table 1 lists some of the big tech investors in leading LLM labs, excluding Google DeepMind since it is fully owned by Google. We believe that it is important to assess the current and future potential for market concentration in this sector due to the increasing importance of foundation models and the already evident impact of concentrated technology markets on economies and societies.

| Company | Big Tech investors |
| --- | --- |
| OpenAI | Microsoft |
| Anthropic | Alphabet, Salesforce, Zoom |
| Inflection AI | Microsoft, Nvidia |
| Hugging Face | Qualcomm, Google, IBM, Intel, Amazon, AMD, NVIDIA, Salesforce |
| Cohere | Nvidia, Oracle, Salesforce |

Table 1: Big tech investors in leading LLM labs. Source: Collected by authors.

## The economics of foundation models

This section describes the technological and economic forces that characterize the market for foundation models. On the supply side, we analyze the technological forces that generate significant economies of scale and certain economies of scope for foundation models. We also observe that certain important inputs to production – data, computational power, and talent – are limited in their supply. On the demand side, we identify network effects that are likely to increase with more economic use. These forces together imply a strong tendency for foundation models towards natural monopoly.

### Supply-side economies of scale

Operating foundation models involves three main types of costs: a significant fixed cost; if they are fine-tuned for specific customers, an additional cost per fine-tuned version; and then a low variable cost to operate. Economies of scale for foundation models arise from the fact that the fixed costs involved in training are so significant. These fixed costs include:

1. The cost of computational power to pre-train the model
2. The cost of acquiring datasets
3. The cost of talent
4. Other costs, including energy, other infrastructure, etc.

Fixed costs usually run into millions of US dollars, driven primarily by the costs of computational power and talent. For instance, the cost of training GPT-4 is estimated to be over USD 100 million (The Economist 2023).

Fine-tuning refers to the process of training a pre-trained model for a specific purpose, usually by using application-specific data. The cost of fine-tuning a foundation model includes:

1. The time of in-house workers
2. The time of outsourced workers, frequently used in the case of Reinforcement Learning based on Human Feedback (RLHF)[7]
3. The computational power required to fine-tune the model

While exact figures are difficult to obtain, these costs are much lower than the fixed cost of pre-training a model because fine-tuning requires less time, data, and computational power. One factor that could push up this cost is the cost of data required for fine-tuning. Fine-tuning requires data that is usually labeled, purpose-relevant, and therefore costlier to obtain. When a foundation model company provides access to the foundation model to an AI application company, the purpose-relevant data is usually provided by the latter. Thus, this cost is borne by the (business) customer rather than the producer, and the customer in most cases tends to already have this data. For instance, if OpenAI provides access to GPT-4 to a healthcare provider, the healthcare provider can use its already existing data on treatment patterns to fine-tune the model to its needs. However, if OpenAI wanted to provide healthcare services itself, it would need to incur the cost of acquiring this data.

The variable cost of inference is more difficult to analyze. Technically, the cost of a single inference is low. But these models are commercially useful only at a certain scale, for instance, only if they respond to millions of queries a day for a given business. One operative question for businesses is how many inferences it takes to replace a human worker, and what the cost of this number of total inferences is.

Since fixed costs are high and variable costs relatively low, foundation models are a classical example of a product that exhibits economies of scale, i.e., the average cost of producing output declines as more output is produced. Economies of scale are a barrier to entry because a new entrant to the market must produce at scale in order to have a competitive cost structure.

---

[7] RLHF is a technique whereby a model's outputs are rated by human evaluators. The model is then able to incorporate this feedback to change or polish its results.

### Network effects or demand-side economies of scale

There are some network effects (also called demand-side economies of scale) for foundation models, but as of now, they are relatively minor in comparison to the economies of scale on the supply side. One source of network effects is that interactions with a generative model can create training data that makes the model better, contributing, for example, to the reinforcement learning that teaches chatbots how to better interact with their users (Serban et al. 2017). OpenAI explicitly states that the data generated by interactions with ChatGPT is collected and may be used for model improvement (Schade, n.d. a).

An additional source of network effects might be ancillary services such as prompt design and prompting tutorials, which are more likely to be geared toward models with the largest market share.

There may also be some demand-side economies of scope that arise from bundling different services. Bundling allows for one-stop shopping by users, lowers search costs, and enhances complementarity between related products (Henten and Windekilde 2022). Companies producing foundation models have already begun bundling the use of their model with various upstream (e.g., computer servers to run the models)[8] and downstream (e.g., integration with business processes through plugins[9]) services, and can do the same for ancillary (e.g., data cleaning) services. Such bundling can act as a barrier to entry by increasing the cost of entry for a new firm.

### Supply-side economies of scope

Economies of scope are significant due to the general-purpose nature of foundation models. A single foundation model, with relatively minor tweaks, can be used for different purposes across industries. For instance, an LLM can be used to automate the creation of holiday itineraries as well as to check for errors in code. These are examples of supply-side economies of scope: using a common capability across different business areas lowers costs overall. Because of the cross-industrial and cross-functional applications made possible by foundation models, the economies of scope are expected to be very large, given that a very large range of services is likely to be affected. OpenAI, for instance, has used plugins to exploit these economies of scope, allowing users to order groceries, search for flights, learn languages, and shop online, all by using the same foundation model (OpenAI 2023b).

---

[8] For instance, companies (including ones that offer open-sourced models) will often earn by providing the computational power to run the model.
[9] See, for instance, OpenAI's ChatGPT plugins https://openai.com/blog/chatgpt-plugins.

In summary, because of these economies of scale on the supply and demand side as well as the economies of scope, the market structure for the most advanced foundation models is likely to be highly concentrated, with a small number of players.[10]

**First-mover advantages**

In the presence of strong economies of scale and scope, the first entrants into a market typically enjoy first-mover advantages that make the market difficult to contest. Suarez and Lanzolla (2005) show that the strength of such first-mover advantages, and by extension the contestability of the market, depend on the pace of technological change within the market and the growth rate of the market itself. It is reasonable to conclude that, as of now, the apparent advances in foundation model technology over the last few years could be due to productization rather than inherent technological improvements. Nevertheless, LLM technology has improved significantly since a decade ago – a pace of change characterized perhaps through peaks rather than rapid and sustained improvements. Additionally, due to economies of scope, market growth in foundation models will likely be fast rather than slow.

In a situation where technological growth is relatively slow and market growth is relatively rapid, we should expect first-mover advantages to not be sustained automatically, but only through special efforts. An incumbent who invests large amounts in productization, marketing, and distribution will be the only one able to sustain these first-mover advantages. The fact that capability jumps occur in a series of peaks means that incumbents have a high incentive to ensure that these jumps happen through their own research, and are brought to market quickly. Indeed, OpenAI seems to be pushing in this direction and its release of ChatGPT seems to have pushed other market participants in the same direction. For instance, after ChatGPT was released, Google changed its long-standing practice of releasing AI papers freely, opting instead to release them only after productization (Tiku and Vynck 2023). On the other hand, Meta has been releasing its models for public use – since it is not the first mover, it has chosen a strategy of attempting to become the primary player in the open-source ecosystem.

We should note that some capability improvements in the future could change these dynamics entirely. If foundation models develop the ability to improve themselves even incrementally, the pace of technological change would be rapid and sustained. This would cause durable first-mover advantages to only accrue to incumbents who make even larger upfront investments in marketing, production, distribution, and research at once. Intellectual property protections, along with access to other inputs like data and computational power, can also help incumbents sustain their position in this scenario.

---

[10] In fact, from a first-best perspective, it would be efficient to have a single producer – training multiple large foundation models to perform the same tasks would duplicate costs unnecessarily. In practice, a single producer may lead to production efficiency but would give that single player substantial market power, which would lead to large inefficiencies from monopoly distortions.

## Access to limited resources

One of the main drivers of first-mover advantages is that the initial entrants to a market have secured access to the production factors and other resources required to operate in that market. Late entrants to the market may find it difficult to access the same resources, may have to pay higher prices, and may face a coordination problem to secure access to complementary resources. Below, we examine three types of limited resources that are important for foundation models today: data, computational power, and talent.

### Data

Foundation models are trained on very large quantities of data. The most data-intensive language model at the time of writing is Google's FLAN, trained on 1.87 trillion words (Epoch 2023). Models are data-hungry, and publicly available data on the internet only contains around 100 trillion words (Epoch 2023). At current rates of growth, it is estimated that AI models may run out of all text on the internet to train on by 2040 and run out of professionally edited text on the internet by 2024 (Villalobos et al. 2022). We should note that models see capability improvements when they are trained on higher-quality text (as demonstrated by the PaLM-2 family of models), and so possessing high-quality texts can be important to differentiate models (Anil et al. 2023).

This makes proprietary datasets very useful for creating or maintaining a competitive advantage. Currently, this advantage lies with Big Tech companies, who control the vast majority of all data generated online – including platform interactions, searches, emails, photos, videos, and other documents. Access to data is therefore an important reason why the producers of foundation models will be interested in vertically integrating with Big Tech companies.

Fine-tuning for certain applications also requires proprietary data – for instance, a shoe manufacturer will be able to derive the most value from a foundation model by fine-tuning it on the manufacturer's historical data. Because specific proprietary data is held in downstream companies, foundation model companies might find it difficult to replace downstream companies entirely. In this respect, they will differ from e-commerce platforms like Amazon, which collect sales data from third-party sellers on that platform and have been accused of using that data to outcompete the same sellers through private labels (Mattioli 2020). In many cases, vertical integration with platforms will allow foundation model companies to overcome this issue. We will see in a later section that foundation models can morph into platforms themselves by being plugged into various applications.

There are emerging techniques that can reduce the cost of acquiring data. These include simulation learning (where a simulated environment substitutes for a real training environment), self-play (where a model can interact with itself to improve its performance), and synthetic data generation (Hwang 2018; Azizi et al. 2023). Some of these techniques are domain-specific and might not work for all foundation models, while other techniques like synthetic data generation come with problems including decreased accuracy. When these techniques are successful, they increase computational power requirements as computation is required to generate synthetic data or learning environments for models. This elevates the importance of computational power as a barrier to entry.

Overall, the need for data and its scarcity will provide incentives for the producers of foundation models to integrate vertically with data controllers. It is likely that large technology companies are already using data collected for other purposes, for example from email services, websites, online documents, or even videos to train foundation models (Victor 2023). Foundation models could potentially be characterized as an attempt to improve products and services, and thus companies can argue that additional consent to use personal data to train models is not required under GDPR[11]-like privacy laws. In its recent order in relation to voice assistant service Alexa, the FTC prohibited Amazon from using ill-gotten children's data to train Alexa's algorithms.[12] This order can serve as an important precedent in preventing AI companies from wantonly interpreting privacy laws and policies to train models on personal data. Moreover, competition authorities may have justifications to prohibit such use of personal data for reasons of restricting vertical integration.

## Computational power

Computational power or "compute" is required to train and run AI models. "Scaling laws" for large models, show that model performance increased predictably with the amount of computational power used (Kaplan et al. 2020). To make better models, developers have to add proportionally more computational power. This is important because computational power is expensive to acquire, has limited supply, and has technical limitations even with innovations in use (Lohn and Musser 2022). While there has been new evidence (Hoffmann et al. 2022) showing that scaling laws might be a little different than was shown in Kaplan et al. (2020), what is not under dispute is the fact that scale is important, and that computational power is a necessary condition to achieve scale.

Computational power is a significant cost lever for building any foundation model as of now. Microsoft's Azure reportedly spent USD 1.2 billion to build a supercomputer to train and run OpenAI's models (Holmes 2023). While these costs are high, they are not high enough to deter entry for long at the current level of model performance. The truly significant entry barrier emerges in the endeavor to build larger, more capable models.

---

[11] General Data Protection Regulation (EU 2016/679)
[12] A press release describing the 2023 order can be found at https://www.ftc.gov/news-events/news/press-releases/2023/05/ftc-doj-charge-amazon-violating-childrens-privacy-law-keeping-kids-alexa-voice-recordings-forever.

We should also note that many estimations of computational power costs usually take into account only the computation required for the final training run, and not the computation required for trial training runs used to arrive at a workable architecture.

Computational power requirements increase with increases in model size. Estimates of the rate of growth of compute requirements vary. Researchers estimate that the computational requirements for training large-scale AI models double roughly every 10 months (Sevilla et al. 2022). The latest projections estimate that the computational cost alone of training a foundation model would exceed US GDP by 2036 if model size increased according to current trends (Lohn and Musser 2022). Other analysts project costs based on model size and find that training a 10-trillion parameter model (one order of magnitude larger than GPT-4) in three months would cost hundreds of billions of US dollars, and power requirements would be extraordinary (Patel 2023).

Recent increases in demand for computational power may drive up costs even further. The costs of manufacturing additional equipment for producing computer chips are high, as they rely on highly specialized manufacturing techniques and high fixed costs (S. M. Khan, Peterson, and Mann 2021). Some of these cost factors are also driven by the fact that the semiconductor design and manufacturing industry – which is the bedrock of computational power provision – is highly concentrated and has several bottlenecks along its supply chain.

There are also forces that pull in the other direction. There are currently strong incentives, driven by returns to investment and government subsidies, for the semiconductor supply chain to become less concentrated. New breakthroughs in computing technology, such as quantum computing, neuromorphic computing, or improvements in memory technology, might potentially ease the supply of computational power in the future.

Overall, the balance currently lies in favor of computational power being a significant barrier to entry to the market of foundation models. Once again, the characteristics of this barrier to entry mean that the foundation models market might be characterized by very few market leaders and intense competition behind the frontier.

**Talent**

Currently, the demand for researchers and engineers who can build frontier foundation models and who can build the server farms necessary for such models far outstrips supply. This is evidenced by the fact that the producers of foundation models are willing to hire engineers who have not worked on AI projects before, even if this means in-house training. The Chicago-based think tank Macro Polo analyzed accepted papers to NeurIPS, the most prestigious AI conference to map frontier AI talent. They found that in 2019, Google (including DeepMind) was the largest single entity with the most NeurIPS papers; Stanford University was a distant second (MacroPolo n.d.).[13]

A McKinsey survey revealed that as of 2021, most organizations found it difficult to hire for AI-related roles (Chui et al. 2022). For instance, 32 percent of all survey respondents found it very difficult to hire AI data scientists, while 46 percent found it somewhat difficult. Interestingly, companies that saw the most returns from AI use found it easier than other organizations to hire AI talent, but they still faced difficulties and attempted to close the gap through upskilling (Chui et al. 2022).

This means that new entrants to the market would, at least for some roles, have to hire people who already work with market leaders. This increases the cost of entry, assuming that a worker generally does not prefer to switch to a job at a company that lags behind their current employer without a significant pay increase. The supply of engineers and researchers is price-inelastic in the short term because it takes a number of years to acquire the expertise necessary to contribute to building foundation models. Additionally, when talented engineers improve algorithmic and software efficiency, they cut down on the exorbitant cost of compute capacity. This drives up demand for these workers, and, in turn, the cost of acquiring such talent

There is a much greater supply of engineers who can build systems behind the frontier – making it likelier that we observe more intense competition behind the frontier. However, innovation in foundation models does not always directly translate into commercial success. For instance, engineers at Google created the Transformer architecture that underpins most foundation models today, but OpenAI was the first to commercialize models based on this architecture (Love and Alba 2023). Competitors may reduce this barrier to entry by adopting the technology developed by market leaders. On the other hand, the increasing secrecy about innovation in frontier foundation models makes this strategy less likely to succeed.

**Intellectual property protections**

Intellectual property (IP) in the form of trade secrets has become a growing entry barrier to the market of foundation models. This was not always the case, and even today, not all developers of foundation models follow this recipe. The Transformer architecture that underpins today's foundation models was developed in a research paper that was publicly released in 2017 (Vaswani et al. 2017); in 2018, Google released an LLM called BERT based on this architecture. BERT is an open-source model, which means that anyone can perform pre-training runs with its architecture, given sufficient computing resources.[14] The astounding performance of BERT on many tasks led to a proliferation of foundation models, many of which were open-source. These include Meta's RoBERTa, OpenAI's [GPT-1](#) and [GPT-2](#), [Stable Diffusion](#), and Google's [LaMDA](#).

---

[13] The researchers at Macro Polo used a "fractional count" to determine the contribution of each entity, assigning a value of 1 to each accepted paper, which was then further divided across all contributing authors. See their methodology at https://macropolo.org/digital-projects/the-global-ai-talent-tracker/methodology-for-global-ai-talent-tracker/.

However, there has been growing secrecy around the technical specifications of frontier foundation models. OpenAI has withheld key technical information about GPT-4, including model size and architecture, justifying this lack of information with both "the competitive landscape" and "safety concerns" (OpenAI 2023c). Similarly, Google has withheld model size and architecture details from the PaLM-2 technical report (Anil et al. 2023). While OpenAI's frontier models are no longer open-source, it reportedly plans on releasing less capable open-source versions of its models, presumably to reap the benefits of developer contribution to its models. DeepMind moved in the opposite direction for the protein-folding model AlphaFold; while the model was initially closed, about a year later the organization released a newer version that was open-source (Hassabis 2021).

AI developers can apply intellectual property protections at all stages of the AI life-cycle, including protections for data, the annotation protocol, labeling, model training, the interface of the inference software, and deployment infrastructure (Gemmo.AI 2022) OpenAI has stated that model outputs belong to the user – which is an important enabling condition for the commercialization of the model (Schade n.d. b).

Counter-intuitively, open-sourcing has also been used by some actors as a tactic to earn greater revenue. Open-sourcing a model incentivizes people to join the "ecosystem" of the AI company, which in turn allows the AI company to extract revenue from adjacent component markets (Ferrandis and Lizarralde 2022). For foundation models, component markets can include hosting services – some open-sourced foundation models are easy to use only with the company's own hosting services. However, this does not necessarily require open-sourcing. OpenAI provides a paid service with dedicated computational capacity and engineering support for its closed models, demonstrating the viability of this component market (Wiggers 2023).

A model that is evaluated more is likely to be more reliable, and an open-source platform also allows commercial users to benchmark model performance. For instance, OpenAI has open-sourced its evaluation platform, allowing people to evaluate its models (OpenAI 2023a).

Open-source alternatives to closed foundation models are still being built. Notable examples are OctoML's [Red Pajama](#), LMSYS.org's [Vicuna](#), and Hugging Face's [BLOOM](#). A serious limiting factor to the success of open-source models outside Big Tech is that as of now, these models do not have the same capabilities as the leading closed models, at least partly due to high costs. It is possible that these models catch up with leading models in some time, but the costs of computational power will still have to be recovered in order for open-source models to scale sustainably.

---

[14] Note that resources include computational power, which can be prohibitively expensive; however, this section examines intellectual property protections as an entry barrier on its own.

Meta has released its latest LLM LlaMA 2 under a license that allows for free commercial use. It is widely believed that this will benefit Meta by promoting the creation of a developer ecosystem around its models, improving their use in Meta's business lines, especially advertising (O'Regan, Victor, and Efrati 2023).

Following this strategy was also suggested in the leaked memo from a Google engineer, which claimed that Google and OpenAI had no competitive "moats," that open-source development could add functionality that their own teams could not, and that Meta would benefit through its focus on nurturing an open-source ecosystem (Patel and Ahmad 2023). The memo claimed that much smaller models such as Alpaca could approach capabilities similar to large models by training on the outputs of frontier models (Alpaca is a small version of Meta's language model LlaMA that was fine-tuned by Stanford researchers on ChatGPT outputs (Taori et al. 2023)). However, subsequent research has shown that such "imitation models" do not approach the capabilities of larger models; although they imitate the performance of larger models for general inquiries, they do not possess the same breadth and depth of capabilities of leading foundation models (Gudibande et al. 2023).

An important insight from the leaked memo, however, is that trade secrets may not be the most significant barrier to entry, and that companies may instead be able to dominate markets by integrating vertically, or "owning their ecosystem".

### The resulting market structure

We have seen that access to talent and computational resources is restricted at the very top – that is, for state-of-the-art models – but not as restricted for models behind the frontier. For talent, we have seen that this is because tacit knowledge is less important behind the frontier. Besides, increasing automation in machine learning engineering itself is likely to reduce the demand for talent behind the frontier[15].

For computational resources, models that perform at lower levels than frontier models are possible to be built with computational resources that too are behind the frontier, the supply of which is more abundant and cheaper, even if they do increase the time taken to train the model. Models behind the frontier also tend to be smaller, requiring less computational power overall.

Currently, the largest and most capable foundation models are created by a few leading AI companies, including OpenAI, Google DeepMind, and Anthropic. Moreover, Meta has recently invested significant resources in foundation models that it has released as open-source models in an attempt to catch up with the market leaders. There are some advanced models that are open-source and created collaboratively, but they do not perform as well as the most capable models. Moreover, several other companies – large companies as well as smaller ones funded by venture capital – have started investing in creating their own smaller foundation models. The market leaders for large models also have smaller or less advanced offerings.

---

[15] See, for instance, DeepMind's AlphaTensor model that automates the discovery of algorithms, including those commonly used in machine learning (Fawzi et al 2022).

For the reasons we examined above, we believe that we are likely to continue to see a concentrated market for the most advanced models, and more competition for smaller, less advanced models. A small number of dominant firms might be able to set the price for the leading models, and other firms might compete for the residual demand for less capable models.

This analysis does not examine the kinds of intermediaries that will emerge in this market, such as companies that perform fine-tuning for other companies, prompt engineers, data brokers, and so on.

## Strategic behavior

Profit-maximizing incumbents may have incentives to increase their monopoly power and establish additional entry barriers. Below are some ways in which foundation model companies can engage in such strategic behavior.

### Vertical integration

Vertical integration refers to the combination of different stages of production under one company or group. On the upside, vertical integration can create its own economies of scale and scope, but on the downside, it can act as a barrier to entry and reduce competition, with all the resulting downsides for consumers. It increases the cost of entry into the market because competitors need to invest in more than one part of the value chain to enter the market with comparable production costs.

Producers of foundation models may engage in vertical integration both upstream and downstream. Upstream vertical integration refers to integration with suppliers, including suppliers of computational power and data. Downstream vertical integration refers to integration with producers of fine-tuned models, and also distributors or producers of final goods and services.

Foundation model production currently exhibits two types of integration with computational power providers. The first type of integration is that controllers of computational power themselves produce foundation models. An example of this is Google's LLM PaLM, which was trained on Google's own chips.[16] For fine-tuning PaLM, computational power is drawn from Google Cloud. In April 2023, Google Brain and DeepMind merged into one team primarily to avoid duplicating their efforts (Google DeepMind 2023).

The second type of integration with computational power providers occurs when foundation model companies and Big Tech companies engage in exclusive contracts. Microsoft's investment in OpenAI gave Microsoft the right to be the exclusive provider of cloud services for OpenAI. This provision covers both training and provision of the model (Warren 2023a). Likewise, Google Cloud is the preferred provider of cloud services for Anthropic (Anthropic 2023). These exclusive and preferred contracts matter in part because we have already started to observe shortages in the market for cloud servers, and exclusive contracts help smooth over such periods of shortage (Holmes and Gardizy 2023). NVIDIA, the dominant chip designer in the frontier AI segment, has been using preferential chip access to prop up new players like CoreWeave in the cloud computing market to make it more competitive, but this just moves the incentive to integrate further upstream (Hjelm n.d.). Microsoft for instance has already signed a deal with CoreWeave to procure NVIDIA's latest GPUs for OpenAI's computing needs (Novet 2023b).

Instances of downstream vertical integration have mushroomed recently. Google and Microsoft have both integrated foundation model capabilities into their products, including Microsoft Office, Gmail, and Google Documents. At the time of writing, Microsoft is charging USD 30 per user per month for its Copilot service that leverages AI in Office apps, Teams, and other productivity tools (Warren 2023b). Likewise, Google is planning to charge business users the exact same amount for its Duet AI assistant that will add generative AI to Google's Gmail and office suite – a potential harbinger of an emerging duopoly (Novet 2023a). Google Search also has a trial version that incorporates generative AI (Reid 2023). These instances can be interpreted as either product improvements or vertical integration. It all depends on whether one considers generative AI services as a natural extension of email and office products or as a separate product that is bundled with email and office, akin to the way Microsoft bundled Internet Explorer with Windows 95. At some level, foundation models are both a product and a feature; the relative prominence of one or the other aspect becomes important if the use of a foundation model in downstream products (like email or office) creates an entry barrier for other potential producers of those downstream products.

This distinction becomes clearer in the case of OpenAI, a company that, until recently, focused on research but decided to provide commercial products to finance the costs of its computational power. OpenAI announced plugins for ChatGPT, allowing the model to be directly linked to commercial applications (OpenAI 2023b). Through its plugins,

---

[16] PaLM was trained on Tensor Processing Units, which are a type of chip developed by Google for use in AI development. For more, see https://ai.googleblog.com/2022/04/pathways-language-model-palm-scaling-to.html.

ChatGPT is now able to, for instance, order groceries from a third-party platform. Insofar as a plugin allows ChatGPT to make choices among providers of groceries, or even to present information about different grocery providers so that the user can choose their preferred provider, plugins allow ChatGPT to function like a platform. It follows that such plugins (or other attempts to turn into a platform) also bring with them all the economic problems presented by platforms, with an added layer of opacity. Sam Altman's recent promise that OpenAI would not compete with its customers underscored the discomfort in the market about the ability of foundation model companies to cannibalize downstream markets (Habib 2023a[17]). These concerns echo the concerns that third-party sellers on Amazon have had, discussed in a later section. In further support of the idea that it may turn into a platform, OpenAI has reportedly considered launching an app store for AI software (Holmes and Victor 2023).

To provide another example of the potential perils of vertical integration, before OpenAI released GPT-4, it gave privileged access to a few select partners that could start building on its foundation before any competitors, giving rise to accusations of cronyism (Thompson 2023). At least one of these partners – Stripe – has also received early-stage investments from OpenAI CEO Sam Altman (Altman 2023, 34:302023). Preferential access has the potential to allow the producers of foundation models or their affiliates to vertically expand their market power without market competition, with all the associated antitrust implications.

Foundation models' integration into platforms, and in some cases morphing into platforms, is likely to present significant competition challenges to markets across the world.

**Pricing strategies**

Market leaders can strategically charge prices below cost to deter entry to the market, a practice called predatory pricing. This strategy usually hinges on being able to charge a premium above the new cost structure once a significant market share has been achieved. Such instances of predatory pricing abound in the platform industry, where venture capital funding has enabled deep discounting of services intending to own the market.

When OpenAI first released ChatGPT in November 2022, it gave users free access, even though OpenAI CEO Sam Altman's statement that computational power probably costs "single-digit cents" per chat.[18] This followed the typical strategy of tech companies to offer their products below cost to acquire users quickly. As of June 2023, ChatGPT API access cost users USD 0.0015 per 1,000 input tokens and USD 0.002 per 1,000 output tokens. It is unclear whether this price covers OpenAI's cost of providing API access. On the other hand, Altman has expressed that OpenAI is working hard to bring down the cost of intelligence (Habib 2023a). Access to chatbots by other foundation model companies, such as Anthropic's [Claude](), DeepMind's [Sparrow](), and Google's [Bard](), is currently free – which is a clear case of selling below cost. These behaviors seem to be a strategic choice to capture market share in advance of competitors' ability to do so.

---

[17] Sam Altman made these comments in a conversation with Raza Habib that was published on his blog at *Humanloop* and later removed at the request of OpenAI (Habib 2023b).
[18] See the tweet at https://twitter.com/sama/status/1599671496636780546.

The described discounting also matters because entire businesses are already being built on the basis of APIs to leading LLMs. If prices were to rise in the future, it is unclear how easily these businesses can shift to using other foundation models, as they may be tied closely to their current provider; it is also unclear how the small number of other foundation model companies can compete now with comparable products, except at these low prices. If pricing below cost is part of a strategy to eliminate competition, it may violate competition law in several jurisdictions. However, the existence of such a strategy is in practice difficult to prove.

### Collusion

Foundation model companies could collude on price, release dates, performance metrics, evaluations, etc. We are not aware of examples of such collusion so far, but foundation model companies might have an incentive to collude in some situations. For instance, foundation model companies could decide to evaluate one another's models and collude to make such evaluations more lenient. Regulators should closely monitor such developments.

On the other hand, companies might also coordinate with one another in a desirable manner to ensure that they do not race to build unsafe AI models (Hua and Belfield 2021). The voluntary commitments entered into by a group of frontier model companies in July 2023, with the backing of the White House, include provisions for technical collaboration for safety (The White House 2023) and led to the formation of the [Frontier Model Forum](), an industry body aiming for safe and responsible AI development (Microsoft 2023). Competition regulators may consider providing clear guidance on the kinds of safety agreements between foundation model companies that are permitted, including agreements with voluntarily determined shared objectives and non-binding commitments. Such guidance must, of course, ensure that there is no risk of increased market concentration or cartelization from these agreements. Competition regulators might need to coordinate with other regulators, in particular those concerned with standard setting, to allow for safety-based coordination between foundation model companies. Competition regulators may also find it useful to revisit safety exemptions for coordination based on new market or technological conditions.

Similar agreements that companies have entered into to lower the adverse environmental impact of their products have sometimes run into trouble with competition law, and thus there is a strong case to be made to avoid such a situation in the case of foundation models – which exhibit the kinds of negative externalities that environmental degradation does (OECD 2021).

It remains to be seen whether the recently formed Frontier Model Forum leans more towards collusion or cooperation. This group consists of OpenAI, Anthropic, Google, and Microsoft. It launched with the stated objectives of advancing AI safety research, identifying best practices for development and deployment, collaborating with policymakers, academia, and civil society, and supporting AI for development (Google 2023).

## Corporate political activity to obtain strategic regulation

Foundation model companies may engage in lobbying or other corporate political activities to bring forth regulation that is beneficial to themselves, at the cost of would-be market entrants, for instance by strengthening entry barriers (e.g., regulations on computational power), or creating new ones (e.g., Safety standards).

Big Tech firms are widely known to lobby for strategic regulation; this practice has continued in the domain of AI. Google, Microsoft, and others have influenced the European Union's regulatory stance on general-purpose AI, demanding regulation on downstream providers rather than on model providers. Big Tech companies have also lobbied the US government in order to indirectly lobby EU decision-makers on the issue (Schyns 2023).

Recently, the CEOs of both [Google](#) (Pelley 2023) and [OpenAI](#) (Kang 2023) have asked for more government regulation of AI. There is a risk that these appeals for more regulations could cement these companies' status in the market, as regulation can be more onerous to comply with for smaller companies and new entrants. Google's CEO Sundar Pichai, for instance, has upheld the EU's GDPR as a template for regulating AI. Some analysts believe that this is because the GDPR has helped Big Tech because it has more capacity than smaller companies to comply with these regulations and that Google desires the same situation in the field of AI (Protalinski 2020).

Regulatory suggestions from OpenAI's researchers have so far included restrictions on the usage of large amounts of computational power and requiring "proof of personhood" to post content (Goldstein et al. 2023). No matter what the intentions are, these suggestions make it relatively easier for established market participants to comply and therefore maintain their market leadership. OpenAI CEO Sam Altman has expressed concerns that other actors may not implement the same safety standards that OpenAI does and argued that they should be imposed via regulation (Mollman 2023). This stated concern for safety is accompanied, on the other hand, by the rapid commercialization of OpenAI's products. Regulators must independently assess these trade-offs between safety and market concentration at the current stage of technological development to arrive at a regulatory strategy.

In the United States, the recent impetus to stay ahead of China in terms of technological capabilities has provided another rhetorical tool for Big Tech lobbying. The timeline of the US-China race for AI shows that many aspects of the race are influenced by Big Tech (AI Now Institute 2023). In addition to Big Tech, this push for supremacy has also created the ground for smaller technology companies to lobby for government favors (Birnbaum 2023).

Overall, a greater compliance burden can entrench monopolies by increasing the costs of production. Large companies usually have a higher ability to afford regulatory compliance burdens since they involve substantial fixed costs. Competition regulators would need to work with other regulators, including data protection authorities, to ensure that the regulation of foundation models walks the fine line between effectiveness and ease of compliance.

## The implications of market concentration in foundation models

Economists widely believe that market concentration poses significant risks to the efficient functioning of markets. The list of negative effects of market concentration includes the following:

### Monopoly distortions

Monopolies or oligopolies that follow the profit motive tend to sell their products at higher prices and lower quantities than what is socially optimal, extracting rents by limiting the economic and social benefits from more widespread use. We should note that these economic considerations apply when the entities in question are profit-maximizing in a concentrated market. In the case of foundation models, it is unclear whether the current market leaders follow this motive.

In addition to price inefficiency, concentrated markets may also give rise to problems with product quality. Since consumers have less choice and cannot easily move to competitors, companies have less incentive to invest in product quality. This may reduce incentives to introduce innovations that benefit consumers, such as privacy safeguards, user-friendly interfaces, or applications.

In some cases, there may also be positive effects from the reduced quantities that come with monopoly distortions: this holds if foundation models generate significant negative externalities, such as bias, safety problems, or unemployment. In those instances, lower output might be beneficial to minimize these externalities.

### Systemic risks from homogenization

A specific example of the quality problems and negative externalities that are part of monopoly distortions are systemic risks that arise from homogenization (Bommasani et al. 2021). Foundation models will likely be integrated into production and delivery processes for goods and services across many sectors of the economy. We can imagine one foundation model in its fine-tuned versions powering decision-making processes in search, market research, customer service, advertising, design, manufacturing, and many more. If foundation models are integrated into a growing number of economic activities, then widespread, cross-industrial applications mean that any errors, vulnerabilities, or failures in a foundation model can threaten a significant amount of economic activity, producing the risk of systemic economic effects. Following is a list of such systemic risks:

a. Biases or errors in the outputs of a foundation model can be inherited by downstream models. These might not be noticed until the model is in use, and its effects might be observed throughout the economy. An example scenario is if an LLM understands demand patterns for consumer goods based on gender stereotypes and consequently leads to over- or under-production of certain goods.
b. Foundation model providers retain the ability to unilaterally deny customers access to the model. As foundation model companies start engaging in downstream economic activities, they may have incentives to deny access to competitors in those activities. If a large number of businesses are built based on access to one foundation model, pulling such access can make all these businesses unviable, providing enormous power to the provider of the foundation model. A withdrawal of access can be catastrophic for businesses built on top of the foundation model.
c. Foundation models could be vulnerable to malicious attacks such as through data poisoning. Data poisoning refers to the practice of tampering with training data in order to influence the outputs of a model. These outputs can then be reflected in the outputs of fine-tuned models as well. Cybersecurity and data auditing methods at the pre-training level can protect against the risk of data poisoning.

All these risks are amplified by the inherent tendency of foundation models towards market concentration, which implies that market forces cannot adequately address these risks. When the market cannot do its job, it is crucial for regulators to step in and ensure that society's interests are safeguarded, and systemic risks are minimized.

Also relevant here is the fact that foundation models have emergent capabilities, i.e., capabilities that they were not specifically trained for and that were discovered after training. For instance, GPT-3 was not trained to write code but turned out to be able to do so. Fine-tuning also increases foundation model capabilities. The existence of emergent capabilities means that systemic risks from foundation models do not have easily identifiable bounds and have the potential to be dramatic.

### Inequality

Foundation models, in their economic implementation, may increase inequality. Following are two channels through which this is likely:

a. Centralization of intellectual functions through automation: Foundation models, like other AI systems, automate tasks generally performed by people. As of now, it appears that several tasks automated by foundation models are done so at a price that is several orders of magnitude cheaper than workers. The unemployment effects of foundation models are beyond the remit of this paper (for a more thorough exposition, see, e.g., Korinek and Juelfs 2023). However, it can be argued that the automation of intellectual functions presents novel problems for competition regulation. Consider a scenario where a foundation model automates nearly all functions of web design. Due to its cost advantage and given enough time, the foundation model can become the only reasonably quick producer at best, and the sole repository of knowledge at worst, of web design. One indication is that the traffic on the developer forum Stack Overflow declined significantly after the release of ChatGPT, shrinking the free online sharing of software development expertise (Carr 2023). We can imagine the same scenario playing out for other functions, such as designing machines or diagnosing certain medical conditions.

    There is no competition problem per se in the automation of intellectual functions, but rather in the centralization of automated intellectual functions. When such centralized automated intellectual functions are also privatized, serious concerns about the distribution of wealth are raised. Competition regulators will have to consider seriously the possibility of AI-driven enclosure of entire intellectual functions.

b. Unequal access: The ability of foundation model companies to deny access to their services also has implications for inequality, apart from the implications for systemic risk outlined earlier. This inequality can be starkly reflected in international inequality, where countries that develop frontier foundation models have a significant advantage in efficient production and distribution systems, which can affect terms of trade.

### Regulatory capture

A concentrated market for foundation models, combined with the widespread application of foundation models, implies high financial stakes for foundation model companies. This could push Big Tech lobbying and the ensuing regulatory capture beyond even current levels, which are already substantial. Regulatory capture reduces the effectiveness of all the policies we discussed above. Historically, private monopolies have often tended to veer towards lower product quality than otherwise, enabled by regulatory capture.

Regulatory capture and economic lock-ins are also reasons to consider ex-ante regulation. Regulators worldwide have recognized that ex-ante regulations might be crucial for emerging technology, because of the pace of development and profound economic implications of such technology.

## AI safety and market concentration

Future highly capable AI systems could cause wide-ranging and perhaps even irreversible harm to society (Anderljung et al. 2023). These concerns include the potential malfunctioning of powerful AI systems as well as the malicious use of such AI systems. AI safety research is particularly concerned about the potential development of artificial general intelligence (AGI), which is the explicit goal of both OpenAI and Google DeepMind. AGI refers to technology that can perform all cognitive tasks that humans can perform, or at least all economically relevant tasks. Many problems could potentially arise from the further advancement of AI if it proceeds in a reckless and unsafe manner, including misuse by unscrupulous actors, catastrophes arising from errors in understanding human preferences, mass unemployment, or even the sidelining of humans in decision-making about humanity's future (Hendrycks et al. 2022).

AI safety risk – the risk of unsafe frontier AI systems being developed and deployed – interacts with the degree of market concentration or competition in several ways, which we explore in the following section.

### Safety risks from competition

Competition in AI development might speed up the development of foundation models to the extent that society might be unable to install guardrails to prevent safety issues from occurring. Competitive pressures may lead AI developers to skimp on safety research and can speed up companies' development of potentially dangerous capabilities in a bid to outdo competitors. In such a scenario, monopolization through a responsible AGI developer is preferable to breakneck competition because it allows society to have enough time to understand and regulate AI development. AI safety researchers have also made the case that a small number of foundation model developers are easier to monitor.

Some current market leaders have explicitly stated a desire to prioritize safety goals, as evidenced by statements released in February and March 2023.[19] If one accords sufficient credibility to these statements, it would then be prudent to allow the current market leaders to lead, lest other players have different goals. But we have seen earlier how for instance the persistence of the profit motive – and thus that of competitive pressures – is a force that risks working in contradiction to these producers' stated goals of prioritizing safety.

---

[19] See OpenAI's statement on preparing for AGI at https://openai.com/blog/planning-for-agi-and-beyond; see Anthropic's document explaining their views on AI safety at https://www.anthropic.com/index/core-views-on-ai-safety.

Additionally, many firms are currently racing to develop competing generative AI offerings, including Meta, which has a chief AI scientist who appears to see most AI safety concerns as overblown (VentureBeat 2023). The open-source release of Meta's Llama model has already enabled programmers to use Llama's published model weights to create novel and targeted adversarial attacks for LLMs that make it possible to circumvent the safety restrictions built into all other leading LLMs (e.g., ChatGPT, Bard, Claude), thereby seriously undermining those organizations' attempts to keep their AI systems safe (Zou et al. 2023). Big Tech firms – Microsoft and Google – have also invested in market participants who have stated that they want to prioritize safety – OpenAI and Anthropic – and these investments may have an influence on strategic firm decisions – possibly diluting safety as a priority.

## Safety risks from market concentration

However, in another scenario, market concentration might increase risks by enabling firms to make progress toward more powerful models more quickly. As we have seen already, the capital cost of building foundation models can be quite large and increases rapidly when higher capabilities are sought. If markets are concentrated and there are just a small number of large players earning monopoly rents, then they will have greater resources at their disposal than if there were a larger number of players in a competitive market with smaller profits. As a result, market concentration may speed up the accumulation of the resources required to make progress and develop more powerful models.

More powerful models may mean increased safety risks (Anderljung et al. 2023): if a frontier AI model can perform a wider variety of tasks and is used in a greater number of applications, then it has more ways to malfunction and act in unexpected or malicious ways, posing risks to public safety (Hendrycks et al. 2022). If this mechanism proves to be important, then one way to ensure greater safety in AI models may be to slow down progress in model capabilities until model safety can be assured.

Market concentration can enable greater scale not only by enabling market leaders to accumulate resources more quickly but also by exerting monopsony power: They might be able to corner the market and pressure the suppliers of computational power and chips as well as the talent to build foundation models.

### AI safety, economic efficiency, and the overall regulatory approach

In summary, a highly competitive market in foundation models seems to carry significant risks for AI safety by promoting a race to the bottom and making monitoring difficult. On the other hand, a high degree of market concentration can cause faster development of capabilities through the monopolization of scarce resources, impeding the cause of AI safety. Further, market concentration increases the risk of regulatory capture, reducing the ability of governments to enforce safety regulations.

Therefore, from both an economic and a safety perspective, it is prudent for governments to adopt a two-pronged regulatory strategy: governments will have to simultaneously ensure that the market for foundation models is contestable and that existing firms are subjected to high standards for safety, akin to the way public utilities are regulated. The recommendations section explains this strategy in greater detail.

### Foundation models, platforms, and general-purpose technologies

It is worth clarifying the relationship between foundation models and platforms, given the vast repository of economic and legal work on platforms and market concentration that has now been developed (Srnicek 2016; L. M. Khan 2017; Gurumurthy et al. 2019). The primary difference between the two is that foundation models are not necessarily multi-sided markets. Both are a type of infrastructure, but foundation models provide insights and automate functions rather than connecting different sides of a market. In this sense, foundation models automate and displace cognitive functions whereas platforms automate and displace marketplaces with their own privatized marketplace.

However, platforms and foundation models have the potential to interact and intensify market concentration. A foundation model can either be plugged into various services to morph into a platform or be vertically integrated with an existing platform. In these cases, the well-known platform entry barriers of network effects and data access (Khan 2017) come into play for foundation models as well and contribute to greater market concentration for foundation models. For example, a directive at Google currently requires the integration of generative AI in all its products that have more than a billion users (Love and Alba 2023). In this case, foundation models (presumably) improve the performance of platform services, which attracts and retains more users and creates more data for foundation models to be trained on.

The difference between the two means that competition-oriented regulatory interventions intended for platforms cannot necessarily be transposed to foundation models wholesale. This includes regulations or regulatory proposals such as interoperability requirements, mandatory data sharing, ranking fairness, and restrictions on exclusive contracts or the use of clients' non-public data. At the same time, the platformization of foundation models means that some of these regulations need to be urgently repurposed for foundation models because the potential market concentration effect of this platformization is higher than before.

## Policy recommendations

The market for foundation models exhibits many peculiarities, including differing barriers to entry for models of different sizes. It inherits some of the distortions caused by digital platforms; it may also fundamentally transform the economy in a way that digital platforms were unable to. Any regulatory strategy for this market needs to respond to these peculiarities. Therefore, we suggest ensuring market contestability and comprehensive regulation to tackle the novel economic challenges presented by foundation models.

### Ensuring that the market for foundation models is contestable

Despite the tendency towards a natural monopoly, the large size of the potential market for foundation models may imply that the market equilibrium will feature more than one market participant. The role of regulators should not be to determine the number of participants but to ensure that the market is contestable in practice. While regulators are unable to change some technological characteristics of this market (such as the requirement of large computational resources), they can target and discourage strategic behavior that increases the concentration in this market beyond what is determined by technological factors.

This section evaluates measures to tackle strategic behavior, including separating different parts of the market. Since first-mover advantages in this sector might depend on success in commercialization, these are not targeted separately.

#### Limiting vertical integration

As we described, the widespread applicability of foundation models implies that vertical integration is a particular threat to competitive markets. Merger review in this area would have to pay particular attention to how any proposed mergers affect the cost of inputs to this market, and how they might privilege some participants in downstream markets to the detriment of others.

Regulatory oversight will also be required even when coordination does not take the form of mergers. For instance, OpenAI's provision of early access to GPT-4 to some companies (including Duolingo, Stripe, and Morgan Stanley) can privilege them in relation to their competitors (Thompson 2023), warranting careful oversight. This is especially true if the recipients of preferential access have financial ties with the producers of foundation models.
Exclusive and preferred use contracts among computational power providers and foundation model companies are rife – take, for instance, Anthropic's deal with Google to use its cloud services, or OpenAI's deal with Microsoft for the same. Big Tech companies can use their own computational power resources at lower than market prices to develop foundation models at cheaper rates than their competitors. This advantage extends further along the value chain – Google uses its own AI chips, Tensor Processing Units, to train its foundation models.

Vertical integration with platforms presents similar concerns. Platforms may corner markets through network effects and data, which may allow them to extract rents. For instance, Amazon charges 30 percent commissions from sellers and competes with sellers in their downstream markets by using their data generated on the platform. Foundation models integrated into platforms, or turning into platforms, may increase the severity of the negative impact of platforms on the competitiveness of the economy. Regulators should be particularly vigilant of such integration.

Antitrust regulators should pay keen attention to acquisitions made by foundation model companies, especially of startups that might compete with them. Governments should equip antitrust regulators with stronger ex-ante powers to stop acquisitions that can be shown to significantly reduce competition. Ex-post measures in digital markets are often too little, too late as important intellectual property can be transferred before a merger or acquisition is undone.

### Investigating pricing strategies

Foundation model companies already engage in deep discounting to provide access to foundation models, but establishing the existence of predatory pricing requires the regulator to prove that deep discounting is part of a strategy to monopolize. While the standards of proof are high, competition regulators can regularly collect information on cost structures from foundation model companies as a first step to discourage predatory pricing.

### Break-ups

If other measures to ensure contestable markets fail, then as a last resort, foundation model companies may need to be broken up into multiple companies on the basis of functions, akin to the way AT&T was broken up.[20] Lina Khan (2017) has elaborated on how principles from banking regulations could be adapted for separation in digital markets. For instance, the Glass-Steagall Act in the United States separated investment banking from commercial banking in order to avoid perverse incentives in commercial banking, and to protect the economy from spillover effects. Similar separation in the AI industry could take various forms, depending on the size of companies, the severity of risks, and the business models in operation. For instance, authorities could require that:

---

[20] See a summary of events in this break-up here: https://www.latimes.com/archives/la-xpm-1995-09-21-fi-48462-story.html.

1. Investors in companies developing frontier foundation models would not be able to invest in upstream domains like computational power or downstream domains like search;
2. Companies that develop frontier foundation models would not be able to exclusively own or control data flows required to train these models; or even,
3. Companies engaging in certain kinds of AI research could not use this research for commercial applications.

**Governance of computational power**

Due to the complexity of the supply chain for computational power, increasing the competitiveness of computational power markets is extraordinarily difficult, yet regulators can still ensure that this market is not further concentrated. Belfield and Hua (2022) provide a list of options that regulators may use to increase the competitiveness of the computational power market. The options they discuss include disallowing some mergers and acquisitions along the computational power supply chain, or seeking remedial action from entities wishing to merge; targeting abuse of dominance, including prices that are too high or too low, bundling, self-preferencing, etc.; and ensuring that horizontal agreements relating to hardware standards are not anti-competitive.

Regulators can also seek separation of some parts of the computational power market, such as by preventing cloud providers from being invested in foundation model developers.[21]

**Governance of data**

Data governance can aid in preventing undesirable concentration in foundation model markets. Existing data protection laws can institute important limits on the exclusive accumulation of data resources required to build large-scale models. This is likely to be an effective lever once data use moves from freely available web data to proprietary datasets. The following principles from data protection law can interact with the goal of fostering competition:

a. Purpose limitation: Various data protection laws require that data only be used for the purpose for which it was collected. Given these laws, regulators could challenge the legal use of proprietary data in foundation models, as this use could fall outside the stated purpose of the data collection. Platforms that collect data might have to collect consent separately for the use of data through foundation models. Seen a little differently, purpose limitations can reduce the economies of scope of foundation models by limiting their legal use cases.
b. Right to an explanation and algorithmic accountability: While these requirements can improve safety, they may also increase companies' compliance burden. Since many of the associated costs are fixed costs, they can be borne more easily by large companies, giving them a competitive advantage. In this way, high compliance costs may exacerbate market concentration.

---

[21] For more on such separations, see Vipra and Myers West (2023).

Antitrust regulators can also mandate that foundation model companies and cloud companies hold certain data in silos, i.e., not share it for other uses (including new models) or with other companies within their corporate groups (Competition and Markets Authority 2020).

The accumulation of vast amounts of proprietary data to train foundation models also creates concerns about competition, as data in these vast amounts is available only to a handful of large technology entities.

**Facilitating the free movement of talent**

Authorities also have a role to play in ensuring that talent can flow freely in the market. Regulators can do this by discouraging or forbidding non-compete clauses in employee contracts as is the case, for instance, in the state of California (Bonta 2022). The FTC has also recently proposed banning non-compete clauses in general (Federal Trade Commission 2023). Regulators can also examine clauses on non-solicitation, confidentiality, and garden leaves to ensure that they do not unduly restrict competition in labor markets.

**Regulating foundation models**

In concentrated markets, competition and the associated market discipline are blunted, making it necessary that regulators ensure that products meet desirable standards and enhance consumer welfare. In the market for foundation models, this is most relevant for powerful frontier models, which will face less competition due to the natural monopoly forces at play in that market.

If their role in the economy grows sufficiently important over time, it may even be desirable to regulate the providers of frontier models akin to public utilities. Due to the inherent safety risks of this technology, societal welfare needs to be safeguarded through safety regulation. Thus, regulators may need to institute a system of mandatory codes of conduct or even licenses to operate foundation model businesses, which can be conditional upon the business following a set of regulations pertaining to safety, prices, and service quality.

**Ensuring a level regulatory playing field between AI and non-AI products and services**

Foundation models are likely to replace and work alongside non-AI providers (including human providers) of different services. The law is often silent on the liability that AI solutions carry when they engage in the real world, especially in sectors where these solutions are not already common. There is no reason why foundation models should be exempt from sectoral regulations including liability, professional licensing, and professional ethics guidelines. If these regulations are derived from various rights of consumers and citizens, they should apply in appropriate ways to foundation models as well.

For instance, governments worldwide have extensive regulations in the education sector, such as around student privacy, non-discrimination, and educational standards. AI solutions used for the classroom should explicitly be subject to the same regulatory standards. Sometimes this will require clarifications or amendments in regulation because it is difficult to apply regulation in the same way to all technologies, or to technology and humans.

There are cases where human workers are penalized for discounting the analysis of an AI solution in their workplace, creating a lopsided liability burden. For instance, nurses in some US hospitals can disregard algorithmic assessment of a patient's diagnosis with doctor approval but face high risks for such disregard as they are penalized for overriding algorithms that turn out to be right. This may lead nurses to err on the side of caution and follow AI solutions even when they know they are wrong in a given instance (Bannon 2023). While these are private penalties upheld by hospital administrations, there has been at least one case where a nurse was held responsible by an arbitrator for a patient's death because she did not override an algorithm. The arbitrator held that she was pressured by hospital policy to follow the algorithm, and thus her employer was directed to pay damages to the patient's family (Bannon 2023).

To avoid situations where humans defer to AI against their better judgment, liability frameworks should be neutral to ensure that technology follows sectoral regulation and not the other way round. AI technology should not be applied in circumstances in which it does not meet regulatory standards.

Such requirements might be onerous and promote more concentration, and the burden to compensate for these effects must be borne by antitrust authorities. Moreover, such requirements would make the deployment of unsafe and substandard systems less financially attractive. It is also possible that enforcing sectoral standards promotes the development of AI systems that are more sector-specific than general, potentially leading to more competition. More importantly, we expect these requirements to protect people against the degradation of product and service standards, and the erosion of consumer rights, due to the use of AI.

### Price and revenue regulations

By restricting quantity, monopolies can increase prices above socially optimal prices. If the producers of foundation models engage in such practices, regulations capping the revenue or price of the monopolist provider of certain services could help. Revenue or price caps can be based on the cost of a given service or on the rate of return. However, establishing such caps is difficult because regulators may not be able to determine the exact cost structure of the monopolist, incentivize monopolists to lower costs, nor determine an appropriate rate of return. If returns are too high, this can cause over-investment and therefore inefficiency; conversely, political pressure to lower price caps can cause under-investment or quality degradation. Moreover, innovation in the field is proceeding so quickly that cost structures are subject to rapid change. Alternatively, a monopolist could be allowed to set prices by themselves subject to regulatory approval.

**Service quality**

A profit-maximizing monopoly has the incentive to offer lower product or service quality compared to what may prevail in a competitive market. Governments should monitor this and include it as a key factor in an appropriate regulatory or licensing framework. There are a few ways in which such requirements can be instituted for foundation models:

1. Monitoring: The government can make it mandatory to disclose certain information about model performance and safety, such as through a system card.[22]
2. Reliability: Like in utility regulation, the government can require that foundation models perform reliably and that outages, if any, are notified and minimized. Foundation model companies have sometimes arbitrarily pulled access to models due to technological issues.[23] A reliability requirement might be useful for businesses that rely on the use of a model and for whom switching to another model is difficult.
3. Privacy: Governments can require that foundation models adhere to high standards of privacy in their development as well as deployment.
4. Setting standards: The government can set and publish minimum standards on performance and safety. Early examples of such standards are the ISO/IEC CD TR 5469 standard on functional safety and AI systems[24] and the ISO/IEC 23894:2023 guidance on risk management.[25]
5. Auditing for safety: The government can audit models (or mandate audits by third parties) to examine their compliance with standards. Algorithmic auditing is a burgeoning field and can be applied to foundation models.

---

[22] System cards are like scorecards representing the results of internal or external audits of an AI model carried out prior to its deployment (Gursoy and Kakadiaris 2022).

[23] In 2019, OpenAI pulled API access to an early iteration of GPT-2 for fear that it would create misinformation. See https://venturebeat.com/ai/openai-releases-curtailed-version-of-gpt-2-language-model/.

[24] International Organization for Standardization (N.d.).
[25] International Organization for Standardization (2023).

6. Liability standards: A regulatory framework that spells out liability standards for foundation models may also be useful, which can include a list of harms that if caused, can lead to license revocation.
7. Interoperability: Interoperability can reduce switching costs (between different foundation models) for businesses and other users. A good model to examine is Open Banking, which is a policy measure that allows interoperability between different banking businesses, creating a playing field that is comparatively more level both for large banks and smaller fintech companies (Gulati-Gilbert and Seamans 2023). Regulators can also attempt to create interoperability in inputs to foundation models, such as data; a data market that is both privacy-enhancing and accessible to all market participants can reduce the forces toward market concentration. Regulators should also be careful to ensure that any infrastructure for interoperability is not controlled by a potential market leader; this would merely entrench their monopoly.

### Access and non-discrimination

In earlier sections, we have made the case that foundation models may provide intellectual infrastructure for a wide range of economic functions. Anyone excluded from their services may be at great economic disadvantage. The government thus has a case for instituting a non-discrimination requirement for access to foundation models. One parallel is that of electricity access, which is sometimes seen as part of other human rights (Tully 2006). In the US, public utility law prohibits undue or unreasonable price discrimination, requiring that similar customers receiving similar services pay the same prices (Henderson and Burns 1989). Section 205(b) of the Federal Power Act of 1920, for example, prohibits undue preference or prejudice to any person, as well as unreasonable differences in rates, charges, service, and facilities.[26] In the EU, Article 10 of Directive 2002/19/EC gives national regulatory authorities power to impose access requirements on communications networks.[27]

When foundation models morph into platforms through plug-ins or app stores, non-discrimination requirements would prevent foundation model companies from privileging their own downstream products and services over other products and services. As Lina Khan (2017) has described, the use of the essential facilities doctrine – compelling a monopolist to provide easy access to competitors in an adjacent market – can be apt in such situations.

### Reducing systemic risks

With further progress in the dissemination of foundation models throughout the economy, the leading foundation models could direct production or business processes in a significant portion of an industry. If an error or other problem emerges in the functioning of such a model, it could grind the industry to a halt, and even cause ripple effects across the economy. Such errors can be difficult to predict. This raises the question of whether there are critical infrastructures or systems that are better kept off-limit for foundation models.

---

[26] See the text of the Act here: https://www.ferc.gov/sites/default/files/2021-04/federal_power_act.pdf
[27] See the text of the Directive here: https://eur-lex.europa.eu/LexUriServ/LexUriServ.do?uri=OJ:L:2002:108:0007:0020:EN:PDF

There is a useful analogy to systemic risks in the financial sector: Regulators can adapt lessons from financial sector regulations to foundation models. In the financial sector, banks and other financial institutions face higher scrutiny when they are of systemic importance.

Large banks across jurisdictions are required to make proportional capital provisions in the event of risks, such that they have a minimum level of capital at all times, as specified in the Basel Standard on risk-based capital requirements (Bank for International Settlements 2023). In the case of frontier foundation models, regulators could require proportional investments in safety research, in order to provision against the most important safety risks.

Banks and companies participating in financial markets are also required to make minimum disclosures to create trust in the market. For foundation model companies, such minimum disclosures can have a wide range depending on the level of concentration in the market and the level of relevant risks posed. Mandatory disclosures could include:
1. Basic model information, such as size, architecture type, the kinds of datasets used, etc.
2. The performance of the model in a given set of evaluations
3. A description of safety precautions taken by the foundation model company
4. Structured access to the model for researchers that is more permissive than access for commercial applications.

## Conclusions

We have seen that due to significant economies of scale, the market for leading foundation models tends toward concentration. Due to the general-purpose nature of foundation model technology, the potential market for foundation models may be the entire economy as the technology advances further. By implication, the risks of market concentration in this sector are unprecedented. As foundation models may increasingly provide the intellectual infrastructure for various economic functions and drive their automation, they present enormous challenges for regulation.

The economics of the foundation model market – including economies of scale and scope, first mover advantages, access to limited resources, and intellectual property protections – shows that the most likely market structure for foundation models is one where there is concentration in the provision of frontier foundation models and competition in the provision of models behind the frontier.

Vertical integration, in particular, with platforms and Big Tech companies, exacerbates the risk of market concentration not only in the foundation models market but in all downstream markets across industries. Integration with platforms, or the morphing of foundation models into platforms, presents regulatory problems that are especially urgent. Other kinds of strategic behavior – predatory pricing, collusion, and lobbying – can increase concentration risks, even for models behind the frontier.

Antitrust authorities and regulators are well-advised to adopt a two-pronged strategy that targets, on the one hand, strategic behavior and abuse of market dominance, and on the other hand, regulations of the leading foundation models to ensure they do not impose AI safety risks and that they meet sufficient quality standards (including safety, privacy, non-discrimination, reliability, and interoperability standards) to maximally contribute to social welfare. Well-designed oversight also requires regulators to have access to information about the cost structures, performance, and contractual arrangements of the leading producers of foundation models.

Finally, it is also important for competition regulators to closely coordinate with other regulators in their jurisdiction, including commerce departments, data protection authorities, and sectoral regulators. Due to the economies of scope of foundation models, coordination with sectoral regulators is important to avoid regulatory overlap and prevent regulatory gaps.

Carr, David F. 2023. "Stack Overflow is ChatGPT Casualty: Traffic Down 14% in March." Similarweb Insights. https://www.similarweb.com/blog/insights/ai-news/stack-overflow-chatgpt/.

Coldewey, Devin. 2019. "OpenAI Shifts from Nonprofit to "capped-Profit" to Attract Capital." *TechCrunch*. March 11, 2019. https://techcrunch.com/2019/03/11/openai-shifts-from-nonprofit-to-capped-profit-to-attract-capital/.

Chui, Michael, Bryce Hall, Helen Mayhew, Alex Singla, and Alex Sukharevsky. 2022. "The State of AI in 2022—and a Half Decade in Review." McKinsey & Company. https://www.mckinsey.com/capabilities/quantumblack/our-insights/the-state-of-ai-in-2022-and-a-half-decade-in-review#/.

Competition and Markets Authority. 2020. "Online Platforms and Digital Advertising Market Study: Final Report." GOV.UK. July 1, 2020. https://www.gov.uk/cma-cases/online-platforms-and-digital-advertising-market-study.

Dettmers, Tim, Artidoro Pagnoni, Ari Holtzman, and Luke Zettlemoyer. 2023. "QLoRA: Efficient Finetuning of Quantized LLMs." arXiv. https://doi.org/10.48550/arXiv.2305.14314.

Dosovitskiy, Alexey, Lucas Beyer, Alexander Kolesnikov, Dirk Weissenborn, Xiaohua Zhai, Thomas Unterthiner, Mostafa Dehghani, et al. 2021. "An Image Is Worth 16x16 Words: Transformers for Image Recognition at Scale." arXiv. http://arxiv.org/abs/2010.11929.

Elias, Jennifer. 2023. "Google's Newest A.I. Model Uses Nearly Five Times More Text Data for Training than Its Predecessor." CNBC. May 16, 2023. https://www.cnbc.com/2023/05/16/googles-palm-2-uses-nearly-five-times-more-text-data-than-predecessor.html.

Eloundou, Tyna, Sam Manning, Pamela Mishkin, and Daniel Rock. 2023. "GPTs Are GPTs: An Early Look at the Labor Market Impact Potential of Large Language Models." arXiv. http://arxiv.org/abs/2303.10130.

Epoch. 2023. "Data Trends." Epoch. April 11, 2023. https://epochai.org/trends.

Fawzi, Alhussein, Matej Balog, Aja Huang, Thomas Hubert, Bernardino Romera-Paredes, Mohammadamin Barekatain, Alexander Novikov, et al. 2022. "Discovering Faster Matrix Multiplication Algorithms with Reinforcement Learning." *Nature* 610 (7930): 47–53. https://doi.org/10.1038/s41586-022-05172-4.

Federal Trade Commission. 2023. "FTC Proposes Rule to Ban Noncompete Clauses, Which Hurt Workers and Harm Competition." Federal Trade Commission. January 4, 2023. https://www.ftc.gov/news-events/news/press-releases/2023/01/ftc-proposes-rule-ban-noncompete-clauses-which-hurt-workers-harm-competition.

Ferrandis, Carlos Muñoz, and Marta Duque Lizarralde. 2022. "Open Sourcing AI: Intellectual Property at the Service of Platform Leadership." *JIPITEC* 13 (3). https://www.jipitec.eu/issues/jipitec-13-3-2022/5557.

Gemmo.AI. 2022. "Intellectual Property in AI." Gemmo.AI. October 24, 2022. https://gemmo.ai/intellectual-property-protection-for-ai/.

Goldstein, Josh A., Girish Sastry, Micah Musser, Renee DiResta, Matthew Gentzel, and Katerina Sedova. 2023. "Generative Language Models and Automated Influence
40

Knight, Will. 2023. "OpenAI's CEO Says the Age of Giant AI Models Is Already Over." *Wired*, April 17, 2023. https://www.wired.com/story/openai-ceo-sam-altman-the-age-of-giant-ai-models-is-already-over/.
Korinek, Anton, and Megan Juelfs. 2023. "Preparing for the (Non-Existent?) Future of Work." Forthcoming in Bullock, Justin et al. (eds.): *The Oxford Handbook of AI Governance*. Oxford, UK: Oxford University Press. https://academic.oup.com/edited-volume/41989.
Le Scao, Teven, Angela Fan, Christopher Akiki, Ellie Pavlick, Suzana Ilić, Daniel Hesslow, Roman Castagné, et al. 2023. "BLOOM: A 176B-Parameter Open-Access Multilingual Language Model." arXiv. http://arxiv.org/abs/2211.05100.
Lohn, Andrew, and Micah Musser. 2022. "AI and Compute: How Much Longer Can Computing Power Drive Artificial Intelligence Progress?" CSET Issue Brief. Center for Security and Emerging Technology, Georgetown University. https://cset.georgetown.edu/publication/ai-and-compute/.
Love, Julia, and Davey Alba. 2023. "Google's Plan to Catch ChatGPT Is to Stuff AI Into Everything." *Bloomberg*, March 8, 2023. https://www.bloomberg.com/news/articles/2023-03-08/chatgpt-success-drives-google-to-put-ai-in-all-its-products.
MacroPolo. n.d. "The Global AI Talent Tracker." MacroPolo. https://macropolo.org/digital-projects/the-global-ai-talent-tracker/.
Mattioli, Dana. 2020. "Amazon Scooped Up Data From Its Own Sellers to Launch Competing Products." *Wall Street Journal*, April 23, 2020, sec. Tech. https://www.wsj.com/articles/amazon-scooped-up-data-from-its-own-sellers-to-launch-competing-products-11587650015.
Meta. 2023. "Meta and Microsoft Introduce the Next Generation of Llama." *Meta* (blog). July 18, 2023. https://about.fb.com/news/2023/07/llama-2/.
Microsoft. 2023. "Microsoft, Anthropic, Google, and OpenAI launch Frontier Model Forum." *Microsoft* (blog). July 26, 2023. https://blogs.microsoft.com/on-the-issues/2023/07/26/anthropic-google-microsoft-openai-launch-frontier-model-forum/.
Mollman, Steve. 2023. "OpenAI CEO Sam Altman Warns That Other A.I. Developers Working on ChatGPT-like Tools Won't Put on Safety Limits—and the Clock Is Ticking." *Fortune*. March 18, 2023. https://fortune.com/2023/03/18/openai-ceo-sam-altman-warns-that-other-ai-developers-working-on-chatgpt-like-tools-wont-put-on-safety-limits-and-clock-is-ticking/.
Novet, Jordan. 2023a. "Google to charge big businesses $30 per user per month for AI in Gmail and work apps." *CNBC*. Aug 29, 2023. https://www.cnbc.com/2023/08/29/google-will-charge-enterprises-30-a-month-for-duet-ai-in-workspace.html.
Novet, Jordan. 2023b. "Microsoft Signs Deal for A.I. Computing Power with Nvidia-Backed CoreWeave That Could Be Worth Billions." *CNBC*. June 1, 2023. https://www.cnbc.com/2023/06/01/microsoft-inks-deal-with-coreweave-to-meet-openai-cloud-demand.html.
43